\documentclass{article}
\usepackage{paper,times}

\usepackage{amsmath,amsfonts,bm}

\def\eqref#1{equation~\ref{#1}}

\def\1{\bm{1}}

\DeclareMathAlphabet{\mathsfit}{\encodingdefault}{\sfdefault}{m}{sl}
\SetMathAlphabet{\mathsfit}{bold}{\encodingdefault}{\sfdefault}{bx}{n}

\usepackage{enumitem}
\usepackage{hyperref}

\usepackage{url}
\usepackage[table]{xcolor}
\usepackage{multirow}
\usepackage{booktabs}
\usepackage{graphicx}
\usepackage{array} 
\usepackage{makecell}
\usepackage{mathtools}
\usepackage{subcaption}
\usepackage{amsthm}
\usepackage{wrapfig}
\usepackage{marvosym}
\definecolor{oursbg}{RGB}{210, 231, 246}
\newcommand{\hl}{\cellcolor{oursbg}}
\usepackage{arydshln}   
\definecolor{densegray}{gray}{0.45}

\newtheorem{definition}{Definition}[section]
\newtheorem{theorem}{Theorem}[section]

\theoremstyle{remark}
\newtheorem{remark}{Remark}[section]
\title{SparseDecoding: Decoding-Aware Pruning for Accurate and Efficient LLM Inference}

\author{
\makebox[\textwidth][c]{%
Qitong Wang$^{1*}$ \hspace{0.7em}
Xinwei Niu$^{1*}$ \hspace{0.7em}
Mingluo Su$^{1}$ \hspace{0.7em}
Shanwei Zhao$^{2}$ \hspace{0.7em}
Shiai Zhu$^{2}$ \hspace{0.7em}
Huan Wang$^{1\text{\Letter}}$%
}\\[1pt]
$^{1}$Westlake University \quad $^{2}$Ant Group \\[1pt]
\url{https://wang-qitong.github.io/SparseDecoding}
}

\usepackage{marvosym}
\begin{document}

\maketitle

{
\let\thefootnote\relax
\footnotetext{$^*$Equal contribution. \quad $\text{\textsuperscript{\Letter}}$Corresponding author: {\tt wanghuan@westlake.edu.cn}}
}

\vspace{-22pt}

\begin{abstract}
The memory-bound nature of the decoding stage of large language model (LLM) inference incurs significant latency. Layer-wise training-free network pruning approaches guided by the Hessian have been a prominent solution to this problem, as pruning reduces the number of nonzero parameters read from memory during decoding. Nevertheless, typical methods in this line compute the Hessian using pre-collected natural sequences, whereas the model is fed self-generated tokens during decoding, creating a distribution shift between the two sequences. The Hessian calculated on the natural sequence is different from that calculated on the generated sequence. We observe that this discrepancy causes the activation distribution during generation to deviate from that used for pruning, further hurting the pruned model performance.
Moreover, most existing LLM pruning methods that bring actual speedup primarily target the sparse matrix-matrix (SpMM) multiplication,  providing limited support for the sparse matrix-vector (SpMV) operations, which dominate decoding. 
To solve these problems, we introduce \textbf{SparseDecoding}, a principled \textit{decoding-aware} pruning framework tailored for accurate and efficient LLM decoding. Specifically, at the algorithmic axis, SparseDecoding constructs calibration matrices from layer-wise activations collected during the dense-model autoregressive generation, excluding prefill, thereby aligning the pruning objective with the decoding activations. At the system axis, we develop an optimized $N{:}M$ sparse matrix-vector kernel with bitmask indexing and fixed-step traversal. 
Substantial empirical results on representative LLMs (Llama-3.1-8B, Llama-3.3-70B, Qwen3-14B / 32B) demonstrate that our method consistently outperforms standard fixed-text calibration on the long-form generation benchmarks while achieving up to $1.48\times$ end-to-end wall-clock decoding speedup on A100 GPUs.
\end{abstract}

\vspace{-10pt}
\section{Introduction}

Large language models (LLMs) are increasingly used for long-output applications such as document drafting, long-form content generation, and code generation \citep{bai2024longwriter,wu2026writingbench,chen2021codex,jain2024livecodebench}. 
The autoregressive decoding process of LLMs is memory-bound and typically dominates end-to-end inference latency \citep{kim2025overfill,fu2024break}.
Generating such long outputs incurs substantial latency, making inference efficiency a practical concern for deployment. 
Weight pruning addresses the inference cost problem of neural networks by zeroing out insignificant weights~\citep{han2015learning}. In terms of LLMs, due to the compute cost and data restrictions, training-free methods are the dominant ones~\citep{frantar-sparsegpt,pmlr-v328-su26a} for sparsifying LLMs, especially those based on the OBS framework~\citep{Hassibi1992OBS}.
\begin{figure}[t]
    \centering

    \begin{subfigure}[t]{0.495\linewidth}
        \centering
        \includegraphics[
            trim={0.2cm 0.2cm 0.23cm 0.24cm},
            clip,
            width=\linewidth
        ]{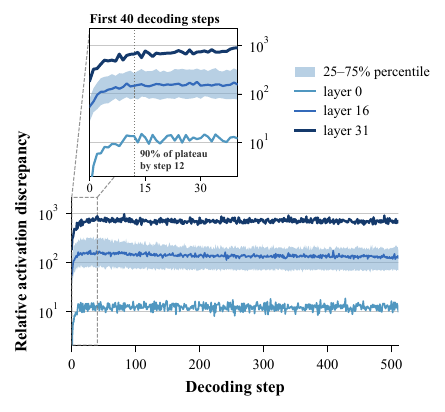}
        \caption{Activation discrepancy during decoding.}
        \label{fig:motivation_drift}
    \end{subfigure}
    \begin{subfigure}[t]{0.495\linewidth}
        \centering
        \includegraphics[
            trim={0.23cm 0 0.2cm 0.1cm},
            clip,
            width=\linewidth
        ]{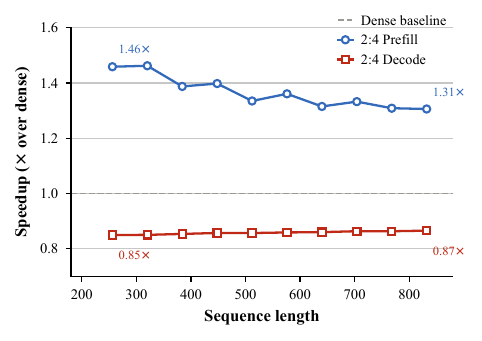}
        \caption{Prefill and decoding speedup.}
        \label{fig:prefill-decode-speedup}
    \end{subfigure}

    \caption{Motivating observations for decoding-aware sparse inference. (a) The relative activation discrepancy between the dense and pruned models grows rapidly during the first few decoding steps before settling into a plateau. The shaded region shows the interquartile range across transformer layers, while the curves correspond to the \textbf{first} (layer 0), \textbf{middle} (layer 16), and \textbf{last} (layer 31) layers. All layers reach 90\% of their plateau by decoding step 12, after which the activation discrepancy remains largely unchanged. (b) Speedup of the $2{:}4$ sparse Tensor Core path over dense execution. Existing sparse kernels provide $1.31$ to $1.46\times$ speedups during the prefill stage, but achieve only $0.85$ to $0.87\times$ of dense performance during autoregressive decoding. Results are obtained on Llama-3.1-8B with \(2{:}4\) sparsity on an NVIDIA A100 GPU.}
    \label{fig:motivation}
    \vspace{-13pt}
\end{figure}

Existing training-free pruning methods~\citep{frantar-sparsegpt,pmlr-v328-su26a} typically rely on a small calibration set to estimate or compensate for the output reconstruction error. Consequently, calibration data play a critical role in determining the quality of the pruned model. Existing methods usually collect calibration activations by running fixed sequences from the calibration set under teacher forcing. This protocol is appropriate for static-input inference settings, where the model inputs remain unchanged by its own predictions. However, autoregressive decoding continuously extends the input context with previously generated tokens. Pruning errors at early decoding steps may therefore alter subsequent predictions and contexts, causing the resulting activation distribution to progressively deviate from that observed during teacher-forced calibration.
As shown in Figure~\ref{fig:motivation_drift}, the relative activation discrepancy increases rapidly during the early stage of autoregressive decoding, with all layers reaching at least \(90\%\) of their eventual plateau within the first 12 decoding steps.
After this initial increase, the discrepancy remains largely stable throughout the subsequent decoding process, even as generation continues for hundreds of additional steps. This behavior suggests that the activation distribution during decoding differs from that observed under fixed teacher-forced calibration, motivating us to construct the calibration objective directly from decode-time activations.

Meanwhile, reducing the theoretical computational cost of pruning alone does not necessarily translate into lower latency, and achieving practical speedups requires efficient sparse-kernel implementations.
We consider $N{:}M$ semi-structured sparsity as a case study, since it offers hardware-friendly regularity while retaining flexibility in selecting weights \citep{fang2024maskllm}. 
Existing $2{:}4$ Sparse Tensor Core libraries, such as cuSPARSELt ~\citep{cusparselt2021} are optimized primarily for Sparse Matrix-Matrix Multiplication (SpMM) \citep{MLSYS2023_a10deb4d}. 
By contrast, operations during decoding are dominated by Sparse Matrix-Vector Multiplication (SpMV) \citep{hong2024flashdecodingfasterlargelanguage}, rather than SpMM, and thus cannot be accelerated by cuSPARSELt implementations.

\begin{figure*}[t]
    \centering
    \includegraphics[trim={0.2cm 0 0cm 0.15cm},clip,width=\linewidth]
    {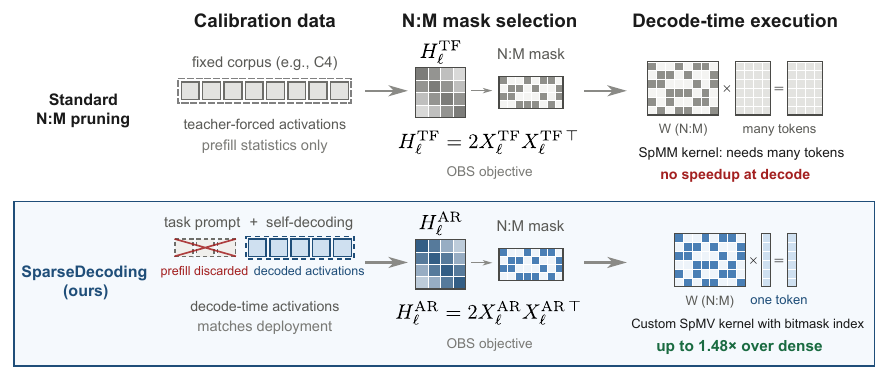}
    \caption{\textbf{Overview of SparseDecoding.} Unlike standard pruning methods collecting calibration activations from fixed corpora, SparseDecoding calibrates on the model-generated tokens from task prompts while discarding the prefill activations. Meanwhile, existing $N{:}M$ sparse Tensor Core kernels accelerate prefill but often underperform dense execution during autoregressive decoding. SparseDecoding instead introduces a decoding-oriented sparse execution pipeline that converts pruned weights into an efficient $N{:}M$ representation and employs an optimized sparse decoding kernel, achieving up to $1.48\times$ decoding speedup under the experimental setup described in Sec.~\ref{sec:experiment}.}
    \label{fig:overview}
\end{figure*}

In this paper, we introduce \textbf{SparseDecoding} (Figure~\ref{fig:overview}) to address these two issues. On the algorithmic side, we build the calibration matrices from layer-wise input activations collected as the dense model generates tokens autoregressively.
We keep only the activations from decoding and discard those from prefill.
This decoding-aware calibration can be used directly with existing pruning methods such as SparseGPT~\citep{frantar-sparsegpt}.
On the system side, inspired by nmSPARSE~\citep{MLSYS2023_a10deb4d}, a GPU library for general \(N{:}M\) sparse computation, we design a customized \(N{:}M\) SpMV kernel for efficient autoregressive decoding. 
Instead of storing a separate 32-bit column index for each nonzero, we represent the sparsity pattern with bitmasks, using one bit per input column and packing every 32 consecutive bits into a 32-bit mask.
For the \(50\%\) \(N{:}M\) patterns with \(M\leq32\) considered in our kernel design, each aligned 32-bit mask word contains exactly 16 set bits.
This allows every output row to traverse its bitmask in the same number of steps, keeping the workload uniform across GPU threads.
Together, decoding-aware calibration and the optimized \(N{:}M\) SpMV kernel preserve generation quality while providing practical end-to-end decoding speedups.
\vspace{10pt}

In summary, this paper makes three contributions:
\begin{enumerate}
    \item We propose a decoding-aware pruning method that aligns the pruning objective with the activations encountered during autoregressive decoding.

    \item We further develop an optimized \(N{:}M\) SpMV kernel for efficient sparse decoding, using bitmask indexing and fixed-step traversal to reduce metadata and indexing overhead.

    \item Extensive experiments across multiple LLMs show that SparseDecoding improves long-form generation quality over existing pruning baselines while achieving up to \(1.48\times\) end-to-end decoding speedup on A100 GPUs.
\end{enumerate}

\vspace{-7pt}
\section{Related Work}
\label{sec:related-work}
\subsection{One-Shot Pruning}
\label{subsec:related-one-shot}
Network pruning reduces the size and inference cost of pretrained models by removing redundant weights~\citep{LeCun1989OBS,han2015learning,Han2015DeepCC,hoefler2021sparsity}. Existing methods can be broadly classified into three categories based on pruning granularity: structured \citep{ma2023llmpruner,10106130}, unstructured \citep{han2015learning,NEURIPS2020_d1ff1ec8}, and semi-structured pruning \citep{fang2024maskllm,mishra2021acceleratingsparsedeepneural,NEURIPS2022_06589ec9}. Structured pruning eliminates entire parameter groups, such as filters or channels \citep{He_2017_ICCV,ma2023llmpruner}, while preserving regular computation that dense hardware and software kernels can handle efficiently~\citep{mishra2021acceleratingsparsedeepneural}, but offering less flexibility in selecting individual weights \citep{hoefler2021sparsity}.
Unstructured pruning provides greater flexibility, but its irregular nonzero locations introduce indirect memory access, limited parallelism, and scheduling overhead \citep{han2015learning,NIPS2016_41bfd20a}.
Semi-structured pruning lies between these two extremes. A representative example is $N{:}M$ sparsity~\citep{NEURIPS2021_6e8404c3}, which retains $N$ nonzero weights in every group of $M$, balancing hardware efficiency with flexible weight selection. 

\subsection{Calibration Data for Pruning}
\label{subsec:related-calibr}
One-shot pruning methods (e.g., SparseGPT \citep{frantar-sparsegpt}, Wanda \citep{sun2023wanda}, ROSE~\citep{pmlr-v328-su26a}) typically use a small calibration set, such as C4~\citep{raffel2020c4} or WikiText~\citep{merity2017pointer}, to estimate weight importance. 
Recent studies have shown that the choice of calibration data can substantially affect pruning quality. 
By comparing multiple pretraining and downstream datasets, \citet{bandari-etal-2024-c4} show that C4 is often not the best calibration source across tasks.
\citet{ji2025beware} further show that self-generated calibration data better aligned with the model's distribution can improve pruning performance.
These studies focus on which calibration corpus to use, whereas SparseDecoding changes how calibration activations are collected specifically from the autoregressive decoding generations.
More closely related to our setting, RAC~\citep{lucas2026reasoning} uses both input activations and activations from the model's on-policy chain-of-thought trajectory for layer-wise reconstruction, while RESP~\citep{wang2025thinkpruneselfreflectivestructured} uses self-generated reasoning traces with a decode-only gradient objective to estimate importance for structured pruning.
Both methods incorporate decode-time information into pruning, while we focus on how the calibration objective differs from the actual decode-time reconstruction objective.
SparseDecoding builds the calibration matrix from task-conditioned decode-step activations, excluding prefill, and uses it to define the layer-wise reconstruction objective. Theorem~\ref{thm:calibration-discrepancy} measures the exact worst-case relative discrepancy between the calibration and decode-time reconstruction objectives.

\subsection{Sparse GPU Kernels and Efficient Decoding}
\label{subsec:related-gpu-kernel}
Whether a sparse model is faster in practice depends on how the sparse pattern is executed on GPUs. 
General sparse kernels such as Sputnik~\citep{sgk_sc2020} show that locality and load balancing can improve sparse matrix computation in deep learning, but irregular sparsity also introduces metadata, indirect memory access, and scheduling overhead. 
NVIDIA cuSPARSELt~\citep{cusparselt2021} targets structured sparse matrix-matrix computation and is more suitable for prefill or larger-batch GEMM/SpMM. In autoregressive decoding, however, the core linear layers at batch one or small batches are closer to GEMV/SpMV, so Tensor-Core-oriented $2{:}4$ paths do not directly cover our target setting. 
\citet{MLSYS2023_a10deb4d} provide a general execution path for $N{:}M$ sparse weights by supporting both SpMV and SpMM. We therefore develop an \(N{:}M\) SpMV execution path specialized for sparse decoding, using bitmask indexing and fixed-step traversal. Our implementation achieves up to a \(1.48\times\) end-to-end decoding speedup on A100 GPUs.

\vspace{-3mm}

\section{Proposed Method: SparseDecoding}
\label{sec:SparseDecoding}
SparseDecoding is a pruning framework for layer-wise post-training pruning and can be combined with OBS-based solvers such as SparseGPT~\citep{frantar-sparsegpt}.
We first introduce the preliminaries on OBS and the relevant works (Sec.~\ref{subsec:preliminary}), and then elaborate on our proposed method (Sec.~\ref{subsec:method-calibration}). Of note, the method is not only for more accurate pruning. We also propose a kernel design (Sec.~\ref{subsec:method-kernel}) to achieve actual wall-clock speedup on hardware.

\subsection{Preliminary}
\label{subsec:preliminary}

\textbf{Layer-wise post-training pruning.}
Directly optimizing the entire model for post-training pruning is computationally prohibitive due to the large number of parameters. Therefore, a widely adopted strategy is to decompose the global compression problem into a sequence of layer-wise reconstruction problems \citep{frantar2022obc,dong2017layerwise}. For layer $\ell$, given the layer-wise calibration activations $X_\ell$ and a layer-wise target sparsity $S_\ell$, the goal is to find a sparse weight matrix $\widehat{W}_\ell$ that changes the layer output as little as possible on $X_\ell$:

\begin{equation}
\widehat{W}_{\ell} = \operatorname*{argmin}_{\widetilde{W}_{\ell}} \left\lVert \left( W_{\ell} - \widetilde{W}_{\ell} \right) X_{\ell} \right\lVert_{F}^{2} \quad \text{s.t.} \quad \operatorname{sparsity}\left(\widetilde{W}_{\ell}\right) = S_{\ell},
\label{eq:layerwise_pruning}
\end{equation}
where $\widetilde{W}_\ell$ ranges over all weight matrices that satisfy the layer-wise sparsity constraint \(S_\ell\), $\widehat{W}_\ell$ denotes the sparse weight matrix that minimizes the objective, and $\left\lVert\cdot\right\rVert_F^2$ denotes the squared Frobenius norm. Solving this problem for each layer sequentially and applying the layer-wise sparse weights yields the final pruned network.

\textbf{Optimal Brain Surgeon for layer-wise pruning.}
The Optimal Brain Surgeon (OBS) framework~\citep{Hassibi1992OBS} efficiently addresses the layer-wise pruning problem in Eq.~(\ref{eq:layerwise_pruning}).
Since the objective is based on the squared $\ell_2$ norm, it decomposes across the rows of $W_{\ell}$ into independent subproblems.
Within each row, OBS uses a second-order Taylor approximation of the reconstruction error, with Hessian $H = 2X_\ell X_\ell^\top$, shared across rows as it depends only on the calibration activations. This approximation admits a closed form for \textbf{(1)} identifying the least salient weight $w_q$ in the row, whose removal induces the smallest increase in reconstruction error; \textbf{(2)} computing the optimal update $\delta w$ to the row's surviving weights that compensates for removing it. The saliency $\mathcal{L}_q$ and the update $\delta w$ are given by
\begin{equation}
\mathcal{L}_q = \frac{w_q^2} {2[H^{-1}]_{qq}}, \quad \delta w = - \frac{w_q} {[H^{-1}]_{qq}} H^{-1}_{:,q}.
\label{eq:obs_update}
\end{equation}
Here, $[H^{-1}]_{qq}$ denotes the $q$-th diagonal entry of the inverse Hessian, and $H^{-1}_{:,q}$ denotes its $q$-th column.
The procedure is applied iteratively, one weight at a time, until the target sparsity $S_\ell$ is reached.
At each pruning step, removing a weight requires the corresponding inverse Hessian information to be updated. 
Performing a full matrix re-inversion after every removal is computationally prohibitive. 
SparseGPT \citep{frantar-sparsegpt} addresses these challenges by fixing the pruning order in advance, enabling efficient, stable maintenance of the required inverse Hessian information.

\subsection{Proposed Method: SparseDecoding}
\label{subsec:method-calibration}
SparseDecoding is a decoding-aware pruning framework that aligns post-training pruning with the autoregressive states encountered during generation.
In OBS~\citep{Hassibi1992OBS}, the calibration data affect which weights are removed and how the remaining weights are updated through the Hessian $H=2X_{\ell}X_{\ell}^{\top}$.
Standard calibration collects $X^{\mathrm{TF}}_\ell$ from fixed text such as C4  \citep{raffel2020c4}, where every token in the context is given in advance. 
Consequently, the Hessian summarizes activations under fixed contexts rather than the model-generated contexts.

However, autoregressive decoding differs from calibration on fixed text because later model states depend on the tokens generated at earlier steps. Errors introduced by pruning at early decoding steps can therefore affect subsequent token predictions and the contexts conditioned on by the model, with these effects accumulating over generation. As a result, the decode-time activation distribution can gradually diverge from that observed under fixed-text calibration.
Thus, we first let $\mathcal{C}=\{q_m\}_{m=1}^{K}$ be a set of calibration prompts drawn from the target task.
Rather than running fixed text through the model, we let the \emph{dense} model
produce its own continuation of each prompt,
\begin{equation}
  y_t^{(m)} \sim P_\theta\big(\cdot \mid q_m,\, y_{<t}^{(m)}\big),
  \qquad t = 1,\ldots,T_m ,
\end{equation}
so that the context at every step is induced by the model's own previously generated outputs.
For prompt $q_m$, we write $\mathcal{P}_m$ for the prompt positions consumed during prefill and $\mathcal{D}_m$ for the $T_m$ positions emitted during decoding.
For a linear layer $\ell$ with weights \(W_\ell\in\mathbb{R}^{d_{\operatorname{out}}\times d_{\operatorname{in}}}\), let \(x_{\ell,t}^{(m)}\in\mathbb{R}^{d_{\operatorname{in}}}\) denote its input activation at step $t$.
We record $x_{\ell,t}^{(m)}$ for every prunable linear layer and keep only the steps $t\in\mathcal{D}_m$, discarding the activations produced while processing $\mathcal{P}_m$.
We discard prefill positions because our calibration objective specifically targets the token-by-token, self-conditioned decoding regime.
As shown in Figure~\ref{fig:motivation_drift}, the activation discrepancy grows rapidly during the early decoding steps, motivating our calibration on decode-time activations.
Then we collect every retained activation to construct a matrix, with each activation being a column of the matrix,
\begin{equation}
  X_\ell^{\mathrm{AR}}
  = \big[\,x_{\ell,t}^{(m)}\,\big]_{m=1,\ldots,K;\; t\in\mathcal{D}_m}
  \in \mathbb{R}^{d_{\operatorname{in}}\times N_{\mathrm{D}}},\quad
  N_{\mathrm{D}} = \sum_{m=1}^{K} |\mathcal{D}_m|,
\end{equation}
where $N_{\mathrm{D}}$ denotes the total number of decoding steps across the $K$ prompts.
During the model pruning phase, we therefore find the sparse weights \(\widehat{W}_\ell\) that follow the minimization objective in Eq.~(\ref{eq:layerwise_pruning}) to minimize the influence on the layer output based on these activations. Written out, our specific autoregressive layer-wise pruning reconstruction objective is defined as:
\begin{align}
\begin{split}
  \mathcal{L}_{\ell}^{\mathrm{AR}}
  &= \big\| ( W_\ell - \widehat{W}_\ell )\, X_\ell^{\mathrm{AR}} \big\|_F^2 \\
  &= \sum_{m=1}^{K} \sum_{t\in\mathcal{D}_m}
     \big\| ( W_\ell - \widehat{W}_\ell )\, x_{\ell,t}^{(m)} \big\|_2^2 ,
\end{split}
\end{align}
so the loss ranges over all $N_{\mathrm{D}}$ generated tokens.
While fixed-text calibration on C4~\citep{raffel2020c4} or reference text constructs \(X_\ell^{\operatorname{TF}}\) from teacher-forced activations, our method instead constructs \(X_\ell^{\operatorname{AR}}\) from autoregressive decode-time activations. The pruning solver itself remains unchanged, but the resulting reconstruction objective is defined over activations that more closely reflect those encountered during generation, without requiring additional training or a new optimizer.

\subsection{N:M SpMV Kernel Design}
\label{subsec:method-kernel}

We first instantiate the framework under $N{:}M$ semi-structured sparsity for our method.
The $N{:}M$ pattern alone does not determine decoding efficiency. 
However, as shown in Figure~\ref{fig:prefill-decode-speedup}, existing $2{:}4$ Sparse Tensor Core kernels reach $1.31$-$1.46\times$ speedup during prefilling but only achieve $0.85$-$0.87\times$ of dense throughput during decoding. In this subsection, we explain why the same sparse weights accelerate the prefilling yet remain inefficient during decoding, and describe the decode-time specialized kernel we develop to address this.

\textbf{Decoding is dominated by SpMV operations.}
In batch-one or small-batch autoregressive decoding, each linear layer processes only the hidden state of the current token, so its computation is closer to General Matrix-Vector Multiplication (GEMV) than large-batch General Matrix-Matrix Multiplication (GEMM). 
After pruning, the corresponding operation becomes $N{:}M$ SpMV.
We therefore develop a decoding-specialized \(N{:}M\) SpMV kernel inspired by
nmSPARSE~\citep{MLSYS2023_a10deb4d}. Our kernel takes advantage of the regular
\(N{:}M\) sparsity pattern to reduce the metadata needed to locate nonzero
weights and to make indices be extracted efficiently:

\textbf{(1) Bitmask indexing.}
We first represent the sparsity pattern of each row with a bitmask, where each bit indicates whether the weight at the corresponding column is retained. Because the nonzero weights are stored in increasing column order, the set bits in the mask correspond to the stored weights in the same order, thus removing the need to store a separate column index or local offset for every nonzero. At \(50\%\) sparsity, this representation uses only \(2\) bits of metadata per nonzero, compared with \(32\) bits for the \texttt{int32} indices used by nmSPARSE and \(8\) bits for a \texttt{uint8} offset representation. The metadata size is therefore reduced by \(16\times\) and \(4\times\), respectively, which is especially useful during batch-one decoding, where SpMV is largely memory-bound.

\textbf{(2) Fixed-step bitmask traversal.}
For a general sparse matrix, different rows may contain different numbers of nonzeros, so traversing their bitmasks requires different numbers of steps. With \(50\%\) \(N{:}M\) sparsity and \(M\leq32\), each aligned \(32\)-bit mask contains exactly \(16\) set bits. The number of traversal steps is therefore fixed for every row and known at compile time. We process each bitmask with a fixed-length loop that the compiler can unroll. At each step, a bit-scan operation finds the least significant set bit, which gives the column of the next nonzero, and that bit is then cleared before continuing. Because each row requires the same number of traversal steps, the workload remains uniform across GPU threads, which allows the \(N{:}M\) bitmask representation to be traversed efficiently during SpMV. We provide further implementation details in Appendix~\ref{appendix:kernel}.

\vspace{-4pt}
\section{Experiments}
\vspace{-4pt}
\label{sec:experiment}
We evaluate SparseDecoding by addressing three questions:
\textbf{(1)} whether autoregressive calibration preserves quality on long writing tasks;
\textbf{(2)} whether the same calibration procedure generalizes to code generation; and
\textbf{(3)} whether the optimized SpMV kernel reduces decoding latency.
Together, these experiments cover the quality and efficiency goals introduced in Sec.~\ref{sec:SparseDecoding}.

\subsection{Experimental Setup}
\vspace{-4pt}
\paragraph{Models.}
We evaluate four open-weight instruction-tuned language models from two model families at two parameter scales: Llama-3.1-8B, Llama-3.3-70B~\citep{grattafiori2024llama3herdmodels}, and Qwen3-14B/32B~\citep{yang2025qwen3technicalreport}. We prune only linear layers supported by the pruning backend, leaving all other components unchanged. The corresponding dense model serves as the quality reference throughout our experiments. We conduct all of our experiments on NVIDIA A100 80 GB GPUs.

\paragraph{Benchmarks.}
Our method is specifically designed for long decoding scenarios. Thus, here we employ two representative benchmarks, \textit{WritingBench}~\citep{wu2026writingbench} and \textit{ClassEval}~\citep{du2023classevalmanuallycraftedbenchmarkevaluating}, to evaluate our method against others:

\textbf{WritingBench for writing.}~\textit{WritingBench}~\citep{wu2026writingbench} contains $1{,}000$ writing prompts belonging to 100 subdomains within 6 domain categories, which are \textit{Academic \& Engineering} \textbf{(D1)}, \textit{Finance \& Business} \textbf{(D2)}, \textit{Politics \& Law} \textbf{(D3)}, \textit{Literature \& Arts} \textbf{(D4)}, \textit{Education} \textbf{(D5)}, and \textit{Advertising \& Marketing} \textbf{(D6)}. Each prompt also introduces three extra requirements as the evaluation criteria: \textit{style, format}, and \textit{length}. Following the original setup, we decode at the temperature of 0.7, top-\texttt{k} value of 20, and top-\texttt{p} value of 0.8, with maximum generation lengths of $16{,}000$ tokens. As the benchmark requires an additional LLM as the evaluation judge, we select DeepSeek-V4-Flash~\citep{deepseekai2026deepseekv4} as the judge. All reported scores are on a scale of 1-10, with \textit{higher} scores indicating better performance.

\textbf{ClassEval for coding.} \textit{ClassEval}~\citep{du2023classevalmanuallycraftedbenchmarkevaluating} contains 100 hand-written Python class generation tasks covering 410 methods, averaging 33.1 tests per class. Given a class skeleton and natural-language description, the model generates the entire class in a single pass. We report class-level Pass@1, the fraction of tasks for which the model's greedy generation passes its class-level tests.
\vspace{-6pt}
\paragraph{Pruning configurations.}
We evaluate $2{:}4$ and $8{:}16$ $N{:}M$ sparsity, as well as $50\%$ unstructured sparsity.  All configurations retain \(50\%\) of the weights, which holds the nominal sparsity constant across the three patterns.  
The two $N{:}M$ settings differ only in the grouping constraint; for instance, a $2{:}4$ mask selects two weights from each group of four, while an $8{:}16$ mask selects eight weights from each group of sixteen. 
Unstructured pruning provides a reference without the local $N{:}M$ constraint. All models we use in our experiments are pruned only once after calibration. 

\begin{table*}[t]
  \caption{\textbf{Results on the WritingBench benchmark under $50\%$ and $2{:}4$ sparsity.}
  {SparseDecoding (ours)} calibrates on autoregressive generations conditioned on LongWriter prompts; C4 uses fixed C4 sequences.
  \textcolor{densegray}{Dense} is the unpruned reference.
  Columns report the overall score, the six domain scores, and the three requirement scores for \textit{style} (R1), \textit{format} (R2), and \textit{length} (R3); ``C'' indicates the corresponding category-specific score.
  The six domains are \textit{Academic \& Engineering} (D1), \textit{Finance \& Business} (D2), \textit{Politics \& Law} (D3), \textit{Literature \& Arts} (D4), \textit{Education} (D5), and \textit{Advertising \& Marketing} (D6).
  All scores are on a scale of 1-10. Bold marks the better overall score between SparseDecoding and C4 within each sparsity setting, based on unrounded values.}
  \label{tab:WritingBench-main}
  \centering
  \small
  \setlength{\tabcolsep}{2.9pt}
  \renewcommand{\arraystretch}{1.2}
  \begin{tabular}{@{}cccc*{12}{c}@{}}
  \toprule
  \multirow{2}{*}{\textbf{Model}} &
  \multirow{2}{*}{\textbf{Sparsity}} &
  \multirow{2}{*}{\textbf{Method}} &
  \multirow{2}{*}{\textbf{Overall}} &
  \multicolumn{6}{c}{\textbf{Domains}} &
  \multicolumn{6}{c}{\textbf{Requirements}} \\
  \cmidrule(lr){5-10}
  \cmidrule(lr){11-16}
  & & &
  & \textbf{D1} & \textbf{D2} & \textbf{D3}
  & \textbf{D4} & \textbf{D5} & \textbf{D6}
  & \textbf{R1} & \textbf{C} & \textbf{R2} & \textbf{C} & \textbf{R3} & \textbf{C} \\
  \midrule

  \multirow{5}{*}{Llama-3.1-8B}
  & \textcolor{densegray}{0\%}
  & \textcolor{densegray}{Dense}
  & \textcolor{densegray}{3.70}
  & \textcolor{densegray}{3.7} & \textcolor{densegray}{3.6}
  & \textcolor{densegray}{3.5} & \textcolor{densegray}{3.3}
  & \textcolor{densegray}{4.1} & \textcolor{densegray}{4.3}
  & \textcolor{densegray}{3.8} & \textcolor{densegray}{3.9}
  & \textcolor{densegray}{3.7} & \textcolor{densegray}{4.3}
  & \textcolor{densegray}{3.8} & \textcolor{densegray}{3.6} \\
  \cdashline{2-16}

  & \multirow{2}{*}{$50\%$}
  & C4
  & 2.25
  & 2.4 & 2.2 & 2.1 & 1.9 & 2.4 & 2.6
  & 2.2 & 2.2 & 2.2 & 2.5 & 2.2 & 2.1 \\
  & &
  \hl\textbf{SparseDecoding}
  & \hl\textbf{2.54}
  & \hl{2.7} & \hl{2.5} & \hl{2.5} & \hl{2.0} & \hl{2.7} & \hl{3.0}
  & \hl{2.5} & \hl{2.5} & \hl{2.5} & \hl{2.9} & \hl{2.5} & \hl{2.4} \\

  \cdashline{2-16}

  & \multirow{2}{*}{$2{:}4$}
  & C4
  & 1.34
  & 1.5 & 1.3 & 1.3 & 1.1 & 1.4 & 1.6
  & 1.3 & 1.3 & 1.3 & 1.4 & 1.3 & 1.2 \\
 & &
  \hl\textbf{SparseDecoding}
  & \hl\textbf{1.52}
  & \hl{1.7} & \hl{1.6} & \hl{1.5} & \hl{1.3} & \hl{1.5} & \hl{1.6}
  & \hl{1.5} & \hl{1.5} & \hl{1.5} & \hl{1.5} & \hl{1.4} & \hl{1.4} \\

  \hline

  \multirow{5}{*}{Llama-3.3-70B}
  & \textcolor{densegray}{0\%}
  & \textcolor{densegray}{Dense}
  & \textcolor{densegray}{4.65}
  & \textcolor{densegray}{4.6} & \textcolor{densegray}{4.5}
  & \textcolor{densegray}{4.5} & \textcolor{densegray}{4.5}
  & \textcolor{densegray}{5.1} & \textcolor{densegray}{5.2}
  & \textcolor{densegray}{4.7} & \textcolor{densegray}{4.9}
  & \textcolor{densegray}{4.6} & \textcolor{densegray}{5.2}
  & \textcolor{densegray}{4.8} & \textcolor{densegray}{4.5} \\
  \cdashline{2-16}

  & \multirow{2}{*}{$50\%$}
  & C4
  & 3.82
  & 3.7 & 3.8 & 3.7 & 3.5 & 4.1 & 4.3
  & 3.8 & 4.0 & 3.8 & 4.3 & 3.8 & 3.7 \\
  & &
  \hl\textbf{SparseDecoding}
  & \hl\textbf{4.01}
  & \hl{4.0} & \hl{3.9} & \hl{3.8} & \hl{3.8} & \hl{4.4} & \hl{4.5}
  & \hl{4.0} & \hl{4.2} & \hl{4.0} & \hl{4.5} & \hl{4.1} & \hl{3.8} \\

  \cdashline{2-16}

  & \multirow{2}{*}{$2{:}4$}
  & C4
  & 3.10
  & 3.1 & 3.2 & 3.0 & 2.6 & 3.3 & 3.5
  & 3.1 & 3.0 & 3.1 & 3.6 & 3.0 & 2.8 \\
  & &
  \hl\textbf{SparseDecoding}
  & \hl\textbf{3.29}
  & \hl{3.2} & \hl{3.3} & \hl{3.0} & \hl{3.1} & \hl{3.7} & \hl{3.8}
  & \hl{3.3} & \hl{3.4} & \hl{3.2} & \hl{3.6} & \hl{3.3} & \hl{2.9} \\

  \hline

  \multirow{5}{*}{Qwen3-14B}
  & \textcolor{densegray}{0\%}
  & \textcolor{densegray}{Dense}
  & \textcolor{densegray}{5.97}
  & \textcolor{densegray}{6.0} & \textcolor{densegray}{5.8}
  & \textcolor{densegray}{6.1} & \textcolor{densegray}{5.7}
  & \textcolor{densegray}{6.3} & \textcolor{densegray}{6.1}
  & \textcolor{densegray}{6.1} & \textcolor{densegray}{6.4}
  & \textcolor{densegray}{6.0} & \textcolor{densegray}{6.8}
  & \textcolor{densegray}{6.0} & \textcolor{densegray}{6.3} \\
  \cdashline{2-16}

  & \multirow{2}{*}{$50\%$}
  & C4
  & 4.62
  & 4.6 & 4.4 & 4.6 & 4.0 & 5.4 & 5.2
  & 4.8 & 5.0 & 4.6 & 5.2 & 4.5 & 4.6 \\
  & &
  \hl\textbf{SparseDecoding}
  & \hl\textbf{5.35}
  & \hl{5.3} & \hl{5.3} & \hl{5.4} & \hl{4.9} & \hl{5.9} & \hl{5.6}
  & \hl{5.5} & \hl{5.8} & \hl{5.4} & \hl{6.2} & \hl{5.4} & \hl{5.6} \\

  \cdashline{2-16}

  & \multirow{2}{*}{$2{:}4$}
  & C4
  & 1.85
  & 2.0 & 1.8 & 1.7 & 1.6 & 2.0 & 2.3
  & 1.9 & 1.9 & 1.8 & 1.7 & 1.9 & 1.7 \\
  & &
  \hl\textbf{SparseDecoding}
  & \hl\textbf{4.18}
  & \hl{4.2} & \hl{4.3} & \hl{4.3} & \hl{3.3} & \hl{4.8} & \hl{4.6}
  & \hl{4.2} & \hl{4.4} & \hl{4.2} & \hl{4.9} & \hl{4.1} & \hl{4.2} \\

  \hline

  \multirow{5}{*}{Qwen3-32B}
  & \textcolor{densegray}{0\%}
  & \textcolor{densegray}{Dense}
  & \textcolor{densegray}{6.48}
  & \textcolor{densegray}{6.5} & \textcolor{densegray}{6.4}
  & \textcolor{densegray}{6.5} & \textcolor{densegray}{6.3}
  & \textcolor{densegray}{6.7} & \textcolor{densegray}{6.6}
  & \textcolor{densegray}{6.6} & \textcolor{densegray}{6.8}
  & \textcolor{densegray}{6.5} & \textcolor{densegray}{7.2}
  & \textcolor{densegray}{6.6} & \textcolor{densegray}{6.8}  \\
  \cdashline{2-16}

  & \multirow{2}{*}{$50\%$}
  & C4
  & 5.34
  & 5.2 & 5.2 & 5.4 & 4.9 & 5.9 & 5.8
  & 5.5 & 5.7 & 5.3 & 6.0 & 5.3 & 5.4 \\
  & &
  \hl\textbf{SparseDecoding}
  & \hl\textbf{6.09}
  & \hl{6.0} & \hl{6.0} & \hl{6.1} & \hl{5.8} & \hl{6.5} & \hl{6.3}
  & \hl{6.2} & \hl{6.5} & \hl{6.1} & \hl{6.8} & \hl{6.1} & \hl{6.4} \\

  \cdashline{2-16}

  & \multirow{2}{*}{$2{:}4$}
  & C4
  & 2.99
  & 3.0 & 2.9 & 2.9 & 2.3 & 3.5 & 3.9
  & 3.0 & 3.1 & 3.0 & 3.3 & 2.9 & 2.8 \\
    & &
  \hl\textbf{SparseDecoding}
  & \hl\textbf{5.31}
  & \hl{5.2} & \hl{5.3} & \hl{5.4} & \hl{4.7} & \hl{5.8} & \hl{5.7}
  & \hl{5.4} & \hl{5.6} & \hl{5.3} & \hl{6.0} & \hl{5.3} & \hl{5.5} \\

  \hline
  \end{tabular}
  \end{table*}
  
\vspace{-4pt}
\paragraph{Calibration configurations.}
The C4 configuration follows the common post-training pruning pipeline. It collects layer inputs from fixed C4 text~\citep{raffel2020c4}. 
SparseDecoding collects layer inputs from decode steps produced by the dense model. The prompts come from LongWriter~\citep{bai2024longwriter} for the writing experiments and from LiveCodeBench problem statements~\citep{jain2024livecodebench} for the code experiments without solutions or reasoning traces. Each dense model generates its own continuation, so the activations we collect are conditioned on the model's own output rather than on reference text.
Both C4 and SparseDecoding use 1M calibration tokens, with only decode-step tokens counted for SparseDecoding. All of the models operate in non-thinking mode during both calibration and evaluation. Both configurations pass their activations to the same SparseGPT backend. Comparisons within one specified sparsity pattern therefore fix the model, pruning backend, and retained weight fraction. We repeat this comparison under both unstructured $50\%$ and semi-structured $2{:}4$ sparsity on Qwen3 models in Sec.~\ref{subsec:ablation}.

\subsection{Main Results: WritingBench}

Table~\ref{tab:WritingBench-main} reports WritingBench scores for the dense checkpoints alongside models pruned with C4 calibration and with SparseDecoding calibration on LongWriter~\citep{bai2024longwriter} prompts.

\paragraph{Effect of decoding-aware calibration.}
SparseDecoding improves the overall WritingBench score across all models and sparsity settings.
Under \(50\%\) unstructured sparsity, the gains range from \(0.19\) to \(0.75\) points across the four models.
Under \(2{:}4\) sparsity, the improvements remain modest for the two Llama models, increasing from \(1.34\) to \(1.52\) for Llama-3.1-8B and from \(3.10\) to \(3.29\) for Llama-3.3-70B.
In contrast, the gains are substantially larger for the Qwen models, with the overall score increasing from \(1.85\) to \(4.18\) for Qwen3-14B and from \(2.99\) to \(5.31\) for Qwen3-32B.
\begin{wraptable}[25]{r}{7.5cm}
\centering
\vspace{-6pt}
\captionsetup{skip=2pt}
\caption{\textbf{Results on the ClassEval benchmark under $50\%$ and $2{:}4$ sparsity.}}
\label{tab:classeval-main}
\footnotesize
\setlength{\tabcolsep}{1.4pt}
\renewcommand{\arraystretch}{1.34}
\begin{tabular}{@{}l c l c@{}}
\toprule
\textbf{Model} & \textbf{Sparsity} & \multicolumn{1}{c}{\textbf{Method}} & \textbf{Pass@1 (\%) $\uparrow$} \\
\hline
\multirow{5}{*}{Llama-3.1-8B}
& \textcolor{densegray}{0\%} & \textcolor{densegray}{Dense} & \textcolor{densegray}{24.0}  \\
\cdashline{2-4}
& \multirow{2}{*}{50\%} & C4 & 2.0  \\
& & \hl \textbf{SparseDecoding} & \hl\textbf{10.0}  \\
\cdashline{2-4}
& \multirow{2}{*}{$2{:}4$} & C4 & 0.0  \\
& & \hl \textbf{SparseDecoding} & \hl\textbf{4.0}  \\
\hline
\multirow{5}{*}{Llama-3.3-70B}
& \textcolor{densegray}{0\%} & \textcolor{densegray}{Dense} & \textcolor{densegray}{33.0} \\
\cdashline{2-4}
& \multirow{2}{*}{50\%} & C4 & 22.0  \\
& & \hl \textbf{SparseDecoding} & \hl\textbf{28.0}  \\
\cdashline{2-4}
& \multirow{2}{*}{$2{:}4$} & C4 & 11.0  \\
& & \hl \textbf{SparseDecoding} & \hl\textbf{15.0}  \\
\hline
\multirow{5}{*}{Qwen3-14B}
& \textcolor{densegray}{0\%} & \textcolor{densegray}{Dense} & \textcolor{densegray}{36.0}  \\
\cdashline{2-4}
& \multirow{2}{*}{50\%} & C4 & 22.0  \\
& & \hl \textbf{SparseDecoding} & \hl\textbf{27.0}  \\
\cdashline{2-4}
& \multirow{2}{*}{$2{:}4$} & C4 & 0.0  \\
& & \hl \textbf{SparseDecoding} & \hl\textbf{6.0}  \\
\hline
\multirow{5}{*}{Qwen3-32B}
& \textcolor{densegray}{0\%} & \textcolor{densegray}{Dense} & \textcolor{densegray}{34.0} \\
\cdashline{2-4}
& \multirow{2}{*}{50\%} & C4 & 25.0 \\
& & \hl \textbf{SparseDecoding} & \hl\textbf{30.0}  \\
\cdashline{2-4}
& \multirow{2}{*}{$2{:}4$} & C4 & 9.0  \\
& & \hl \textbf{SparseDecoding} & \hl\textbf{24.0}  \\
\hline
\end{tabular}
\end{wraptable}
We also observe similar improvements across most writing domains and the requirement scores. Since each comparison fixes the model, pruning backend, retained weight fraction, and sparsity pattern, the results suggest that calibration activations play an important role in preserving generation quality after pruning.

\subsection{Main Results: ClassEval}

Table~\ref{tab:classeval-main} reports class-level Pass@1 for the dense checkpoints alongside models pruned with C4 calibration and SparseDecoding calibration on LiveCodeBench~\citep{jain2024livecodebench} problem statements.

\paragraph{Comparisons over calibration methods.}
SparseDecoding is ahead across all eight model-sparsity pairs by 4 to 15 points.
Comparing with C4 under $2{:}4$ sparsity, Pass@1 rises from $9.0$ to $24.0\%$ on Qwen3-32B, from $0.0$ to $6.0\%$ on Qwen3-14B, from $11.0$ to $15.0\%$ on Llama-3.3-70B, and from $0.0$ to $4.0\%$ on Llama-3.1-8B.
Notably, under 2:4 sparsity, C4 calibration yields $0.0$ on both Llama-3.1-8B and Qwen3-14B, whereas SparseDecoding recovers nonzero scores.

\subsection{Empirical Decoding Speedup}
\label{subsec:speedup}

Table~\ref{tab:e2e-speedup} reports decoding throughput and speedup relative to the dense model across $N{:}M$ patterns from $2{:}4$ to $16{:}32$.
All measurements use NVIDIA A100 GPUs and the GPT-Fast framework~\citep{gptfast}.
On each model, all four patterns achieve nearly identical throughput, with speedups of $1.42\times$ on Llama-3.1-8B, $1.48\times$ on Llama-3.3-70B, $1.35\times$ on Qwen3-14B, and $1.45\times$ on Qwen3-32B. 
This is expected from our kernel design. Since all patterns retain 50\% of the weights and use the same bitmask metadata, they incur identical memory traffic per row. In contrast to the $2{:}4$ Sparse Tensor Core path in Figure~\ref{fig:prefill-decode-speedup}, which runs slower than dense execution during decoding, our kernel turns every evaluated $N{:}M$ pattern into a practical decoding speedup.
\vspace{10pt}

\begin{table*}[t]
\caption{\textbf{End-to-end decoding throughput and speedup over the dense baseline}.
All of the $N{:}M$ patterns retain $50\%$ of the weights, so differences reflect kernel efficiency rather than compression.
Every evaluated pattern decodes faster than the dense baseline, with speedups ranging from \(1.34\) to \(1.48\times\).
Measurements use a context length of 512 and report the median after 50 warmup and 50 measured iterations on NVIDIA A100 GPUs with GPT-Fast.}
\label{tab:e2e-speedup}
\centering
\small
\setlength{\tabcolsep}{7pt}
\renewcommand{\arraystretch}{1.1}
\begin{tabular}{l *{8}{c}}
\toprule
\multirow{2}{*}{$N{:}M$}
& \multicolumn{2}{c}{Llama-3.1-8B}
& \multicolumn{2}{c}{Llama-3.3-70B}
& \multicolumn{2}{c}{Qwen3-14B}
& \multicolumn{2}{c}{Qwen3-32B} \\
\cmidrule(lr){2-3}\cmidrule(lr){4-5}\cmidrule(lr){6-7}\cmidrule(lr){8-9}
& Token/s & Speedup & Token/s & Speedup & Token/s & Speedup & Token/s & Speedup \\
\midrule
\textcolor{densegray}{Dense}
& \textcolor{densegray}{96.26} & \textcolor{densegray}{1.00$\times$}
& \textcolor{densegray}{21.67} & \textcolor{densegray}{1.00$\times$}
& \textcolor{densegray}{39.76} & \textcolor{densegray}{1.00$\times$}
& \textcolor{densegray}{19.15} & \textcolor{densegray}{1.00$\times$} \\
\cdashline{1-9}
\addlinespace[2pt]
$2{:}4$    & 136.30 & 1.42$\times$ & 30.17 & 1.39$\times$ & 53.18 & 1.34$\times$ & 27.69 & 1.45$\times$ \\
$4{:}8$    & 136.43 & 1.42$\times$ & 31.64 & 1.46$\times$ & 53.86 & 1.35$\times$ & 27.83 & 1.45$\times$ \\

$8{:}16$
& 136.89 & 1.42$\times$
& 29.90 &  1.38$\times$
& 53.42 &  1.34$\times$
& 27.77 &  1.45$\times$ \\

$16{:}32$  & 136.89 & 1.42$\times$ & 32.13 & 1.48$\times$ & 53.73 & 1.35$\times$ & 27.81 & 1.45$\times$ \\
\bottomrule
\end{tabular}
\vspace{-10pt}
\end{table*}

\subsection{Ablation Study}
\label{subsec:ablation}

\begin{wraptable}[16]{r}{7cm}
\vspace{-14pt}
\caption{\textbf{Results on the WritingBench benchmark under self-generated and cross-model calibration.} Qwen3-14B is pruned using activations from its own generations \textbf{(self-generated)} or from Qwen3-32B generations \textbf{(cross-model)}.}
\label{tab:ablation-policy}
\centering
\small
\setlength{\tabcolsep}{2.7pt}
\renewcommand{\arraystretch}{1.3}
\begin{tabular}{@{}c c c c@{}}
\toprule
\textbf{Model} & \textbf{Sparsity} & \textbf{Calibration} & \textbf{Overall} \\
\midrule
\multirow{4}{*}{Qwen3-14B}
& \multirow{2}{*}{$50\%$}
& Cross-model (32B)
& {5.23} \\
& &
\hl\textbf{Self-generated (14B)}
& \hl{\textbf{5.35}} \\

\cdashline{2-4}
& \multirow{2}{*}{$2{:}4$}
& Cross-model (32B)
& {4.13} \\
& &
\hl\textbf{Self-generated (14B)}
& \hl{\textbf{4.18}} \\
\hline
\end{tabular}
\end{wraptable}
\vspace{-4pt}
\paragraph{Ablation over calibration context source (Table~\ref{tab:ablation-policy}).}

We examine whether calibration tokens must come from the model being pruned (self-generated) or can be produced by a different model (cross-model). 
Specifically, we compare calibrating the 14B model on its own tokens with using tokens generated by the larger 32B model when pruning the 14B model. 
Self-generated calibration performs better in both settings, achieving $4.18$ vs. $4.13$ under \(2{:}4\) sparsity and $5.35$ vs. $5.23$ under \(50\%\) unstructured sparsity.
This suggests that generating calibration tokens with the model being pruned is a contributing factor, and using sequences generated by a different model introduces a distributional discrepancy in the calibration data.

\paragraph{Varying the pruning backend (Table~\ref{tab:ablation-wanda24}).}
\begin{wraptable}[13]{r}{7cm}
\vspace{-55pt}
\caption{\textbf{Results on the WritingBench benchmark under 50\% and 2:4 sparsity with Wanda on the Qwen3 model family.}}
\label{tab:ablation-wanda24}
\centering
\small
\setlength{\tabcolsep}{4pt}
\renewcommand{\arraystretch}{1.3}
\begin{tabular}{@{}c c c c*{9}{c}@{}}
\toprule
\multirow{2}{*}{\textbf{Model}} &
\multirow{2}{*}{\textbf{Sparsity}} &
\multirow{2}{*}{\textbf{Method}} &
\multirow{2}{*}{\textbf{Overall}} \\
& & & \\

\midrule

\multirow{4}{*}{Qwen3-14B}
& \multirow{2}{*}{$50\%$}
& C4
& {4.94}
\\
& &
\hl\textbf{SparseDecoding}
& \hl{\textbf{5.23}}\\

\cdashline{2-13}
& \multirow{2}{*}{$2{:}4$}
& C4
& {2.41}\\
& &
\hl\textbf{SparseDecoding}
& \hl{\textbf{3.30}}\\

\hline

\multirow{4}{*}{Qwen3-32B}
& \multirow{2}{*}{$50\%$}
& C4
& {5.52}\\
& &
\hl\textbf{SparseDecoding}
& \hl{\textbf{5.71}}\\

\cdashline{2-13}
& \multirow{2}{*}{$2{:}4$}
& C4
& {3.34}\\
& &
\hl\textbf{SparseDecoding}
& \hl{\textbf{4.39}}\\

\hline
\end{tabular}
\end{wraptable}

All preceding experiments use SparseGPT~\citep{frantar-sparsegpt} as the pruning backend when contrasting C4 with SparseDecoding, so the reported gains could in principle be specific to the solver. 
To verify the improvements are robust to the choice of pruning backend, we present the comparison with Wanda~\citep{sun2023wanda} in Table~\ref{tab:ablation-wanda24}. SparseDecoding remains consistently ahead of C4 under both sparsity settings, indicating that improvements come from the calibration activations themselves rather than from the pruning backend.

\vspace{-6pt}
\section{Conclusion}
\vspace{-6pt}
This work presents \textit{SparseDecoding}, a decoding-aware pruning framework for LLMs that bridges the gap between pruning calibration objectives and downstream decoding behavior. 
By incorporating autoregressive outputs into calibration, SparseDecoding improves generation quality without additional training. Additionally, a \(N{:}M\) SpMV kernel is introduced to achieve practical speedup with the weight sparsity.
Extensive experiments across multiple LLMs and sparsity patterns demonstrate that SparseDecoding consistently outperforms conventional calibration approaches, highlighting the importance of decoding-aware calibration for training-free LLM pruning. On top of the method, our \(N{:}M\) SpMV kernel achieves up to $1.48\times$ end-to-end decoding speedup on NVIDIA A100 GPUs.

\bibliography{paper}
\bibliographystyle{paper}

\appendix

\section{Analysis of Our Method}
\label{appendix:analysis}
In this section, we theoretically characterize the discrepancy between a calibration objective and the decode-time reconstruction objective, and empirically compare this discrepancy for autoregressive and fixed-text calibration.

\subsection{Theoretical Analysis}
\label{appendix:theor-analy}

\textbf{Setup.}
We follow the notation introduced in Sec.~\ref{sec:SparseDecoding}, and recall that \(\Delta W_\ell:= W_\ell-\widehat W_\ell\) denotes the weight difference between the dense weights and the sparsified weights of layer \(\ell\).
For an activation source \(R\in\{\operatorname{Dec},\operatorname{AR},\operatorname{TF}\}\), \(\operatorname{Dec}\) denotes decode-step activations produced by the dense model during autoregressive generation on held-out target-task prompts, \(\operatorname{AR}\) denotes decode-step activations produced by the dense model on the calibration prompts used by SparseDecoding, and \(\operatorname{TF}\) denotes activations obtained from fixed calibration sequences.
We let \(x_\ell^R\in\mathbb{R}^{d_{\ell,\operatorname{in}}}\) denote a random input activation of layer \(\ell\) drawn from source \(R\), and define the associated population Hessian as 
\begin{equation}
    H_\ell^R := 2\,\mathbb{E}\left[x_\ell^R (x_\ell^R)^\top \right]\in \mathbb{R}^{d_{\ell,\operatorname{in}}\times d_{\ell,\operatorname{in}}}
\end{equation}

Since \(H_\ell^{\operatorname{Dec}}\) is symmetric positive semidefinite, let
\(\{u_{\ell,i}\}_{i=1}^{d_{\ell, \operatorname{in}}}\) be an orthonormal eigenbasis satisfying
\begin{equation}
    H_\ell^{\operatorname{Dec}}u_{\ell,i}=\lambda_{\ell,i}u_{\ell,i},\qquad\lambda_{\ell,1}\geq    \cdots\geq\lambda_{\ell,d_{\ell, \operatorname{in}}} \geq 0.\label{eq:dec-eigendecomposition}
\end{equation}
For any \(k\) such that \(\lambda_{\ell,k}>0\), we define
\begin{equation}
    U_{\ell,k} := [u_{\ell,1},\ldots,u_{\ell,k}], \quad 
    \Lambda_{\ell,k} := \operatorname{diag} (\lambda_{\ell,1},\ldots,\lambda_{\ell,k}), \quad
    P_{\ell,k} := U_{\ell,k}U_{\ell,k}^{\top}.
    \label{eq:topk-projection}
\end{equation}
Here,
\(U_{\ell,k}\in\mathbb{R}^{d_{\ell, \operatorname{in}}\times k}\) contains the top-\(k\) eigenvectors of \(H_\ell^{\operatorname{Dec}}\), \(\Lambda_{\ell,k}\in\mathbb{R}^{k\times k}\) contains their corresponding eigenvalues, and \(P_{\ell,k}\in\mathbb{R}^{d_{\ell, \operatorname{in}}\times d_{\ell, \operatorname{in}}}\) is the orthogonal projector onto the span of the eigenvectors.

\begin{definition}
\textnormal{\textbf{(Normalized decode-time calibration discrepancy).}}
For a calibration activation source \(C\in\{\operatorname{AR},\operatorname{TF}\}\), we define
\begin{equation}
\widetilde H_{\ell,C}^{(k)} := \Lambda_{\ell,k}^{-1/2} U_{\ell,k}^{\top} H_\ell^C U_{\ell,k} \Lambda_{\ell,k}^{-1/2} \in \mathbb{R}^{k\times k}, 
\label{eq:normalized-hessian}
\end{equation}
and
\begin{equation}
\epsilon_{\ell,C}^{(k)} := \left\lVert \widetilde H_{\ell,C}^{(k)} - I_k \right\rVert_{\operatorname{op}}. 
\label{eq:epsilon-def}
\end{equation}
where \(\widetilde H_{\ell,C}^{(k)}\) is the calibration Hessian \(H_\ell^C\) restricted to the top-$k$ decode-time eigenvectors and rescaled by the corresponding decode-time eigenvalues. The scalar \(\epsilon_{\ell, C}^{(k)}\) is the operator (spectral) norm of the difference between the normalized calibration Hessian and the identity matrix \(I_k\).
\end{definition}

\begin{remark}
We remark that the identity matrix \(I_k\) is the normalized form of the decode-time Hessian, since
\begin{equation}
\Lambda_{\ell,k}^{-1/2}
U_{\ell,k}^{\top}
H_\ell^{\operatorname{Dec}}
U_{\ell,k}
\Lambda_{\ell,k}^{-1/2}
=
I_k.
\end{equation}
Thus, \(\epsilon_{\ell,C}^{(k)}=0\) indicates that the calibration Hessian exactly matches the decode-time Hessian after projection onto the top-$k$ decode-time eigenvectors and normalization by the corresponding decode-time eigenvalues. More generally, a smaller \(\epsilon_{\ell,C}^{(k)}\) indicates a smaller worst-case relative discrepancy between the calibration and decode-time reconstruction objectives over these directions.
\end{remark}

\begin{definition}
\textnormal{\textbf{(Projected reconstruction loss).}}
For \(R\in\{\operatorname{Dec},\operatorname{AR},\operatorname{TF}\}\), we define the layer-wise reconstruction loss over the top-\(k\) eigenvectors of the decode-time Hessian as
\begin{equation}
\begin{aligned}
\mathcal L_{\ell,R}^{(k)}(\Delta W_\ell) &:= \mathbb{E}\left[ \left\lVert\Delta W_\ell P_{\ell,k} x_\ell^R \right\rVert_2^2 \right] \\ &= \frac{1}{2} \operatorname{tr} \left(\Delta W_\ell     P_{\ell,k}H_\ell^R P_{\ell,k} \Delta W_\ell^\top \right).
\end{aligned}
\label{eq:projected-recon-loss}
\end{equation}
\end{definition}

\begin{theorem}
\label{thm:calibration-discrepancy}
\textnormal{\textbf{(Calibration reconstruction-loss discrepancy).}}
For every \(C\in\{\operatorname{AR},\operatorname{TF}\}\),
\begin{equation}
\epsilon_{\ell,C}^{(k)} = \max_{\Delta W_\ell: \mathcal L_{\ell,\operatorname{Dec}}^{(k)}(\Delta W_\ell)>0} \frac{ \left| \mathcal L_{\ell,C}^{(k)}(\Delta W_\ell) - \mathcal L_{\ell,\operatorname{Dec}}^{(k)}(\Delta W_\ell) \right| }{ \mathcal L_{\ell,\operatorname{Dec}}^{(k)}(\Delta W_\ell)}.
\label{eq:relative-recon-error}
\end{equation}
\begin{proof}
We first fix \(C\in\{\operatorname{AR},\operatorname{TF}\}\), and then define
\(Z := \Delta W_\ell U_{\ell,k}\Lambda_{\ell,k}^{1/2}\) and \( M_C := \widetilde H_{\ell,C}^{(k)}-I_k\) to simplify the notation in the remainder of the proof. We first begin our proof by subtracting the two losses
\begin{align}
\mathcal L_{\ell,C}^{(k)}(\Delta W_\ell) - \mathcal L_{\ell,\operatorname{Dec}}^{(k)}(\Delta W_\ell)
&= \frac{1}{2}\operatorname{tr}\left(Z\widetilde H_{\ell,C}^{(k)}Z^\top\right) - \frac{1}{2}\operatorname{tr}\left(ZZ^\top\right)
&& \text{(Definitions of two losses} \text{ and } Z\text{)}\nonumber\\
&= \frac{1}{2}\operatorname{tr}\left(Z\widetilde H_{\ell,C}^{(k)}Z^\top - ZZ^\top\right)
&& \text{(linearity of trace)}\\
&= \frac{1}{2}\operatorname{tr}\left(Z\left(\widetilde H_{\ell,C}^{(k)} - I_k\right)Z^\top\right)\nonumber\\
&= \frac{1}{2}\operatorname{tr}\left(ZM_CZ^\top\right)
&& \text{(Definition of } M_C\text{)}\label{eq:trace}
\end{align}
Since \(M_C\) is symmetric,
\begin{align}
\left\lvert \mathcal L_{\ell,C}^{(k)}(\Delta W_\ell) - \mathcal L_{\ell,\operatorname{Dec}}^{(k)}(\Delta W_\ell) \right\rvert 
&= \frac{1}{2} \left\lvert \operatorname{tr}(ZM_CZ^\top) \right\rvert \nonumber \quad &&(\textnormal{Eq.}~\ref{eq:trace})\\ 
&= \frac{1}{2}\left\lvert\langle Z, ZM_C\rangle_F\right\rvert \quad &&(\textnormal{Definition of } \langle \cdot, \cdot \rangle_F)\\
&\leq \frac{1}{2}\left\lVert Z\right\rVert_F \left\lVert Z M_C\right\rVert_F\quad &&(\textnormal{Cauchy-Schwarz inequality})\\
&\leq \frac{1}{2} \left\lVert Z\right\rVert_F^2 \left\lVert M_C \right\rVert_{\operatorname{op}} \quad \nonumber\\ 
&= \epsilon_{\ell,C}^{(k)} \mathcal L_{\ell,\operatorname{Dec}}^{(k)}(\Delta W_\ell). 
\quad && \text{(Definitions of \(Z, M_C\), Eqs.~\ref{eq:epsilon-def}, \ref{eq:projected-recon-loss})}
\label{eq:proof-upper-bound}
\end{align}
where \(\langle\cdot, \cdot\rangle_F\) denotes the Frobenius inner product, which proves the upper bound in Eq.~(\ref{eq:relative-recon-error}) by dividing by \(\mathcal{L}_{\ell, \operatorname{Dec}}^{(k)}(\Delta W_\ell)\). 

To show that the bound in Eq.~(\ref{eq:proof-upper-bound}) is tight, we construct a specific \(\Delta W_\ell\) that attains the upper bound. Since \(M_C\) is symmetric, it admits an eigendecomposition with real eigenvalues and orthonormal eigenvectors. We let \(v\in\mathbb{R}^k\) be an eigenvector of \(M_C\) whose eigenvalue \(\lambda^*\) has the largest absolute value so that \(|\lambda^*| = \|M_C\|_{\operatorname{op}}\). Then we let \(Z^*\in\mathbb{R}^{d_{\ell,\operatorname{out}}\times k}\) be the rank-one matrix in which the first row is set to \(v^\top\), while the remaining rows are zero. Then the following conditions hold:
\begin{equation}
\|Z^*\|_F^2 = \|v\|_2^2 = 1, \quad 
\left|\operatorname{tr}(Z^*M_CZ^{*\top})\right| = \left|v^\top M_C v\right| = |\lambda^*| = \|M_C\|_{\operatorname{op}}
\label{eq:proof-tightness}
\end{equation}
As the top-$k$ eigenvectors are orthonormal, then the conditions \(\lambda_{\ell,k}>0\) and \(U_{\ell,k}^\top U_{\ell,k}=I_k\) hold, we therefore define \(\Delta W_\ell^* := Z^*\Lambda_{\ell,k}^{-1/2}U_{\ell,k}^{\top} \nonumber\) which recovers \(Z^*\) under the substitution used throughout the proof by \(\Delta W_\ell^* U_{\ell,k}\Lambda_{\ell,k}^{1/2} = Z^*\Lambda_{\ell,k}^{-1/2}U_{\ell,k}^\top U_{\ell,k}\Lambda_{\ell,k}^{1/2} = Z^*\). In particular, we can derive the reconstruction objective
\begin{equation}
\mathcal L_{\ell,\operatorname{Dec}}^{(k)}(\Delta W_\ell^*) = \frac{1}{2}\|Z^*\|_F^2 = \frac{1}{2} > 0, \nonumber
\end{equation}
hence \(\Delta W_\ell^*\) belongs to the feasible set of the maximization in Eq.~(\ref{eq:relative-recon-error}). Substituting into Eq.~(\ref{eq:proof-tightness}) gives
\begin{equation}
\frac{\left|\mathcal L_{\ell,C}^{(k)}(\Delta W_\ell^*) - \mathcal L_{\ell,\operatorname{Dec}}^{(k)}(\Delta W_\ell^*)\right|}{\mathcal L_{\ell,\operatorname{Dec}}^{(k)}(\Delta W_\ell^*)} 
= \frac{\frac{1}{2}\left|\operatorname{tr}(Z^*M_CZ^{*\top})\right|}{\frac{1}{2}\|Z^*\|_F^2} 
= \|M_C\|_{\operatorname{op}} = \epsilon_{\ell,C}^{(k)}, \nonumber
\end{equation}
so the upper bound is achieved exactly at \(\Delta W_\ell^*\), completing the proof of Eq.~(\ref{eq:relative-recon-error}).
\end{proof}
\end{theorem}

\begin{remark}
We remark that by Theorem~\ref{thm:calibration-discrepancy}, 
for a layer \(\ell\) and retained dimension \(k\), if
\begin{equation}
    \epsilon_{\ell,\operatorname{AR}}^{(k)} < \epsilon_{\ell,\operatorname{TF}}^{(k)},
    \nonumber\end{equation}
then $\operatorname{AR}$ calibration has a smaller worst-case relative reconstruction-loss discrepancy with respect to the decode-time objective than $\operatorname{TF}$ calibration at layer $\ell$.
\label{remark:calibration-comparison}
\end{remark}

\subsection{Empirical Findings}
\label{appendix:empir-analy}
In this section, we use prompts from LongWriter~\citep{bai2024longwriter} to generate the autoregressive calibration activations. We instantiate the fixed calibration source $\operatorname{TF}$ using the C4 dataset~\citep{raffel2020c4}, hence, we denote the fixed calibration source as $\operatorname{C4}$ within this section.
We compute the finite-sample version of the calibration discrepancy in Eq.~(\ref{eq:epsilon-def}) for Qwen3-14B and 32B~\citep{yang2025qwen3technicalreport}. For each retained fraction \(\rho\in\{0.01,0.02,0.05,0.10,0.20,0.50\}\), we set \(k_\ell(\rho)=\max\{1,\lceil\rho d_{\ell,\operatorname{in}}\rceil\}\) as the number of top-$k$ eigenvectors to retain.
Let \(\widehat H_\ell^C\) and \(\widehat H_\ell^{\operatorname{Dec}}\) denote the empirical Hessians constructed from collected activation matrices. More generally, for \(R\in\{\operatorname{Dec},\operatorname{AR},\operatorname{TF}\}\), we let 
\(X_\ell^R=[x_{\ell,1}^R,\ldots,x_{\ell,N_R}^R]\in\mathbb{R}^{d_{\ell,\operatorname{in}}\times N_R}\) denote the collected activation matrix where \(N_R\) denotes the number of collected activation samples from the source \(R\). We therefore estimate the corresponding population Hessian \(H_\ell^R\) by
\begin{equation}
    \widehat H_\ell^R
    :=
    \frac{2}{N_R}
    X_\ell^R (X_\ell^R)^\top
    =
    \frac{2}{N_R}
    \sum_{i=1}^{N_R}
    x_{\ell,i}^R (x_{\ell,i}^R)^\top.
\end{equation}

We therefore obtain \(\widehat U_{\ell,k}\) and \(\widehat\Lambda_{\ell,k}\) from \(\widehat H_\ell^{\operatorname{Dec}}\) by Eq.~(\ref{eq:topk-projection}), and define
\begin{equation}
    \hat{\epsilon}_{\ell, C}^{(k)} := \left\lVert \widehat\Lambda_{\ell, k}^{-1/2}\,\widehat U_{\ell, k}^\top\, \widehat H_\ell^C\, \widehat U_{\ell, k}\,\widehat\Lambda_{\ell, k}^{-1/2} - I_k\right\rVert_{\operatorname{op}}
    \label{eq:empirical-epsilon}
\end{equation}
as the finite-sample estimate of \(\epsilon_{\ell,C}^{(k)}\) in Eq.~(\ref{eq:epsilon-def}), obtained by substituting empirical Hessians for the population Hessians.
By Theorem~\ref{thm:calibration-discrepancy}, its population counterpart \(\epsilon_{\ell,C}^{(k)}\) equals the worst-case relative reconstruction loss discrepancy. We therefore use \(\hat\epsilon_{\ell,C}^{(k)}\) as an empirical estimate of this quantity.
By Remark~\ref{remark:calibration-comparison}, a smaller value indicates a smaller empirical worst-case reconstruction-loss discrepancy.
The median ratio of the estimates is reported as
\(\hat{\epsilon}_{\operatorname{C4}}/\hat{\epsilon}_{\operatorname{AR}}\), where \(\hat{\epsilon}_{\operatorname{C4}}/\hat{\epsilon}_{\operatorname{AR}} > 1\) indicates that AR has a smaller calibration discrepancy than C4. Table~\ref{tab:delta} and \ref{tab:delta-32} report the empirical statistics of Qwen3-14B and 32B, respectively. Across both models and all retained fractions, AR calibration yields a smaller discrepancy in at least $87\%$ of the prunable modules, with median ratios $\hat{\epsilon}_{\operatorname{C4}}/\hat{\epsilon}_{\operatorname{AR}}$ between $1.81$ and $3.26$, indicating that decode-time calibration consistently aligns more closely with the decode-time reconstruction objective than fixed-text calibration.

\begin{table*}[t]
\caption{\textbf{Empirical comparison of AR and C4 calibration discrepancy on Qwen3 14B.}
For each retained eigenvector fraction, we report the number and percentage of prunable modules (out of 280) where \(\hat{\epsilon}_{\ell,\operatorname{AR}}^{(k)} < \hat{\epsilon}_{\ell,\operatorname{C4}}^{(k)}\), together with the median of the discrepancy ratio \(\hat{\epsilon}_{\operatorname{C4}}/\hat{\epsilon}_{\operatorname{AR}}\).}
\label{tab:delta}
\centering
\small
\setlength{\tabcolsep}{6pt}
\renewcommand{\arraystretch}{1.1}
\begin{tabular}{cccc}
\toprule
Retained eigenvectors
&
Modules with
\(\hat{\epsilon}_{\operatorname{AR}} < \hat{\epsilon}_{\operatorname{C4}}\)
&
Percentage
&
Median ratio
\(\hat{\epsilon}_{\operatorname{C4}}/\hat{\epsilon}_{\operatorname{AR}}\)
\\
\midrule
1\%  & 274 & 97.9\% & 2.30 \\
2\%  & 271 & 96.8\% & 2.39 \\
5\%  & 266 & 95.0\% & 2.65 \\
10\% & 264 & 94.3\% & 2.83 \\
20\% & 261 & 93.2\% & 3.06 \\
50\% & 248 & 88.6\% & 3.26 \\
\bottomrule
\end{tabular}
\end{table*}

\begin{table*}[t]
\caption{\textbf{Empirical comparison of AR and C4 calibration discrepancy on Qwen3 32B.}
For each retained eigenvector fraction, we report the number and percentage of prunable modules (out of 448) where \(\hat{\epsilon}_{\ell,\operatorname{AR}}^{(k)} < \hat{\epsilon}_{\ell,\operatorname{C4}}^{(k)}\), together with the median of the discrepancy ratio \(\hat{\epsilon}_{\operatorname{C4}}/\hat{\epsilon}_{\operatorname{AR}}\).}
\label{tab:delta-32}
\centering
\small
\setlength{\tabcolsep}{6pt}
\renewcommand{\arraystretch}{1.1}
\begin{tabular}{cccc}
\toprule
Retained eigenvectors
&
Modules with
\(\hat{\epsilon}_{\operatorname{AR}} < \hat{\epsilon}_{\operatorname{C4}}\)
&
Percentage
&
Median ratio
\(\hat{\epsilon}_{\operatorname{C4}}/\hat{\epsilon}_{\operatorname{AR}}\)
\\
\midrule
1\%  & 390 & 87.1\% & 1.81 \\
2\%  & 419 & 93.5\% & 1.92 \\
5\%  & 440 & 98.2\% & 2.08 \\
10\% & 439 & 98.0\% & 2.16 \\
20\% & 444 & 99.1\% & 2.17 \\
50\% & 442 & 98.7\% & 2.19 \\
\bottomrule
\end{tabular}
\end{table*}

\section{Additional Experimental Results}
\label{appendix:more-res}

\paragraph{SparseDecoding for 8:16 sparsity (Table~\ref{tab:ablation-816}).} 
We additionally evaluate SparseDecoding under the \(8{:}16\) sparsity setting. Compared with C4 calibration, SparseDecoding consistently improves generation quality across the evaluated models. These results show that the benefit of decoding-aware calibration is not limited to \(2{:}4\) sparsity, but also extends to the \(8{:}16\) pattern.

\begin{table*}[t]
\caption{\textbf{Results on WritingBench for Qwen3-14B and Qwen3-32B under unstructured $50\%$ and semi-structured $8{:}16$ sparsity.} \textbf{SparseDecoding (ours)} calibrates on autoregressive generations conditioned on LongWriter prompts; C4 uses fixed C4 sequences. \textcolor{densegray}{Dense} is the unpruned reference. ``C'' indicates the corresponding category-specific score. Bold marks the best overall score within each sparsity setting.}
\label{tab:ablation-816}
\centering
\small
\setlength{\tabcolsep}{3pt}
\renewcommand{\arraystretch}{1.15}
\begin{tabular}{@{}cccc*{12}{c}@{}}
\toprule
\multirow{2}{*}{\textbf{Model}} &
\multirow{2}{*}{\textbf{Sparsity}} &
\multirow{2}{*}{\textbf{Method}} &
\multirow{2}{*}{\textbf{Overall}} &
\multicolumn{6}{c}{\textbf{Domains}} &
\multicolumn{6}{c}{\textbf{Requirements}} \\
\cmidrule(lr){5-10}
\cmidrule(lr){11-16}
& & & &
\textbf{D1} & \textbf{D2} & \textbf{D3} &
\textbf{D4} & \textbf{D5} & \textbf{D6} &
\textbf{R1} & \textbf{C} & \textbf{R2} & \textbf{C} & \textbf{R3} & \textbf{C} \\
\hline

\multirow{5}{*}{Qwen3-14B}
& \textcolor{densegray}{$0\%$}
& \textcolor{densegray}{Dense}
& \textcolor{densegray}{5.97}
& \textcolor{densegray}{6.0} & \textcolor{densegray}{5.8}
& \textcolor{densegray}{6.1} & \textcolor{densegray}{5.7}
& \textcolor{densegray}{6.3} & \textcolor{densegray}{6.1}
& \textcolor{densegray}{6.1} & \textcolor{densegray}{6.4}
& \textcolor{densegray}{6.0} & \textcolor{densegray}{6.8}
& \textcolor{densegray}{6.0} & \textcolor{densegray}{6.3} \\
\cdashline{2-16}

& \multirow{2}{*}{$50\%$}
& C4
& 4.62
& 4.6 & 4.4 & 4.6 & 4.0 & 5.4 & 5.2
& 4.8 & 5.0 & 4.6 & 5.2 & 4.5 & 4.6 \\
& & \hl\textbf{SparseDecoding}
& \hl\textbf{5.35}
& \hl{5.3} & \hl{5.3} & \hl{5.4} & \hl{4.9} & \hl{5.9} & \hl{5.6}
& \hl{5.5} & \hl{5.8} & \hl{5.4} & \hl{6.2} & \hl{5.4} & \hl{5.6} \\

\cdashline{2-16}

& \multirow{2}{*}{$8{:}16$}
& C4
& 4.50
& 4.5 & 4.4 & 4.5 & 3.8 & 5.0 & 5.2
& 4.6 & 4.8 & 4.5 & 5.0 & 4.3 & 4.2 \\
& & \hl\textbf{SparseDecoding}
& \hl\textbf{5.31}
& \hl{5.2} & \hl{5.3} & \hl{5.3} & \hl{4.7} & \hl{5.9} & \hl{5.7}
& \hl{5.4} & \hl{5.6} & \hl{5.4} & \hl{6.1} & \hl{5.2} & \hl{5.4} \\

\hline

  \multirow{5}{*}{Qwen3-32B}
  & \textcolor{densegray}{0\%}
  & \textcolor{densegray}{Dense}
  & \textcolor{densegray}{6.48}
  & \textcolor{densegray}{6.5} & \textcolor{densegray}{6.4}
  & \textcolor{densegray}{6.5} & \textcolor{densegray}{6.3}
  & \textcolor{densegray}{6.7} & \textcolor{densegray}{6.6}
  & \textcolor{densegray}{6.6} & \textcolor{densegray}{6.8}
  & \textcolor{densegray}{6.5} & \textcolor{densegray}{7.2}
  & \textcolor{densegray}{6.6} & \textcolor{densegray}{6.8} \\
\cdashline{2-16}

& \multirow{2}{*}{$50\%$}
& C4
& 5.34
& 5.2 & 5.2 & 5.4 & 4.9 & 5.9 & 5.8
& 5.5 & 5.7 & 5.3 & 6.0 & 5.3 & 5.4 \\
  & &
  \hl\textbf{SparseDecoding}
  & \hl\textbf{6.09}
  & \hl{6.0} & \hl{6.0} & \hl{6.1} & \hl{5.8} & \hl{6.5} & \hl{6.3}
  & \hl{6.2} & \hl{6.5} & \hl{6.1} & \hl{6.8} & \hl{6.1} & \hl{6.4} \\

\cdashline{2-16}

& \multirow{2}{*}{$8{:}16$}
& C4
& 5.20
& 5.2 & 5.1 & 5.3 & 4.5 & 5.6 & 5.7
& 5.3 & 5.5 & 5.2 & 5.8 & 5.1 & 5.2 \\
& & \hl\textbf{SparseDecoding}
& \hl\textbf{5.91}
& \hl{5.8} & \hl{5.9} & \hl{5.9} & \hl{5.5} & \hl{6.3} & \hl{6.3}
& \hl{6.0} & \hl{6.3} & \hl{5.9} & \hl{6.7} & \hl{6.0} & \hl{6.1} \\

\hline
\end{tabular}
\end{table*}

\begin{table*}[t]
\caption{\textbf{Results on WritingBench benchmark under $50\%$ and $2{:}4$ sparsity with different calibration data for the Qwen3 model family.} Pile and RedPajama use fixed pretraining-corpus sequences; \textbf{SparseDecoding} calibrates on autoregressive generations conditioned on LongWriter prompts.}
\label{tab:ablation-calib-source}
\centering
\small
\setlength{\tabcolsep}{3pt}
\renewcommand{\arraystretch}{1.15}
\begin{tabular}{@{}c c c c*{12}{c}@{}}
\toprule
\multirow{2}{*}{\textbf{Model}} &
\multirow{2}{*}{\textbf{Sparsity}} &
\multirow{2}{*}{\textbf{Calibration}} &
\multirow{2}{*}{\textbf{Overall}} &
\multicolumn{6}{c}{\textbf{Domains}} &
\multicolumn{6}{c}{\textbf{Requirements}} \\
\cmidrule(lr){5-10}
\cmidrule(lr){11-16}
& & &
& \textbf{D1} & \textbf{D2} & \textbf{D3}
& \textbf{D4} & \textbf{D5} & \textbf{D6}
& \textbf{R1} & \textbf{C} & \textbf{R2} & \textbf{C} & \textbf{R3} & \textbf{C} \\
\midrule

\multirow{7}{*}{Qwen3-14B}
& \textcolor{densegray}{$0\%$}
& \textcolor{densegray}{Dense}
& \textcolor{densegray}{5.97}
& \textcolor{densegray}{6.0} & \textcolor{densegray}{5.8}
& \textcolor{densegray}{6.1} & \textcolor{densegray}{5.7}
& \textcolor{densegray}{6.3} & \textcolor{densegray}{6.1}
& \textcolor{densegray}{6.1} & \textcolor{densegray}{6.4}
& \textcolor{densegray}{6.0} & \textcolor{densegray}{6.8}
& \textcolor{densegray}{6.0} & \textcolor{densegray}{6.3} \\
\cdashline{2-16}

& \multirow{3}{*}{$50\%$}
& Pile
& 5.00
& 5.2 & 5.1 & 5.1 & 4.0 & 5.7 & 5.3
& 5.0 & 5.2 & 5.0 & 5.6 & 4.8 & 5.0 \\
& & RedPajama
& 5.04
& 5.0 & 5.0 & 5.1 & 4.4 & 5.6 & 5.6
& 5.2 & 5.4 & 5.1 & 5.7 & 4.9 & 5.0 \\
& & \hl\textbf{SparseDecoding}
& \hl\textbf{5.35}
& \hl{5.3} & \hl{5.3} & \hl{5.4} & \hl{4.9} & \hl{5.9} & \hl{5.6}
& \hl{5.5} & \hl{5.8} & \hl{5.4} & \hl{6.2} & \hl{5.4} & \hl{5.6} \\

\cdashline{2-16}

& \multirow{3}{*}{$2{:}4$}
& Pile
& 2.68
& 3.0 & 2.7 & 2.6 & 1.9 & 3.0 & 3.1
& 2.7 & 2.7 & 2.7 & 2.9 & 2.4 & 2.2 \\
& & RedPajama
& 2.62
& 2.8 & 2.6 & 2.5 & 2.1 & 2.7 & 3.3
& 2.7 & 2.7 & 2.6 & 2.7 & 2.4 & 2.2 \\
& & \hl\textbf{SparseDecoding}
& \hl\textbf{4.18}
& \hl{4.2} & \hl{4.3} & \hl{4.3} & \hl{3.3} & \hl{4.8} & \hl{4.6}
& \hl{4.2} & \hl{4.4} & \hl{4.2} & \hl{4.9} & \hl{4.1} & \hl{4.2} \\

\hline

  \multirow{7}{*}{Qwen3-32B}
  & \textcolor{densegray}{0\%}
  & \textcolor{densegray}{Dense}
  & \textcolor{densegray}{6.48}
  & \textcolor{densegray}{6.5} & \textcolor{densegray}{6.4}
  & \textcolor{densegray}{6.5} & \textcolor{densegray}{6.3}
  & \textcolor{densegray}{6.7} & \textcolor{densegray}{6.6}
  & \textcolor{densegray}{6.6} & \textcolor{densegray}{6.8}
  & \textcolor{densegray}{6.5} & \textcolor{densegray}{7.2}
  & \textcolor{densegray}{6.6} & \textcolor{densegray}{6.8} \\
\cdashline{2-16}

& \multirow{3}{*}{$50\%$}
& Pile
& 5.80
& 5.8 & 5.8 & 5.8 & 5.3 & 6.2 & 6.2
& 5.9 & 6.1 & 5.8 & 6.5 & 5.8 & 6.0 \\
& & RedPajama
& 5.72
& 5.7 & 5.7 & 5.8 & 5.3 & 6.1 & 6.1
& 5.8 & 6.1 & 5.7 & 6.4 & 5.7 & 5.7 \\
  & &
  \hl\textbf{SparseDecoding}
  & \hl\textbf{6.09}
  & \hl{6.0} & \hl{6.0} & \hl{6.1} & \hl{5.8} & \hl{6.5} & \hl{6.3}
  & \hl{6.2} & \hl{6.5} & \hl{6.1} & \hl{6.8} & \hl{6.1} & \hl{6.4} \\
\cdashline{2-16}

& \multirow{3}{*}{$2{:}4$}
& Pile
& 3.32
& 3.6 & 3.3 & 3.3 & 2.6 & 3.6 & 3.9
& 3.4 & 3.4 & 3.3 & 3.6 & 3.2 & 3.2 \\
& & RedPajama
& 3.74
& 3.8 & 3.7 & 3.7 & 3.0 & 4.2 & 4.5
& 3.8 & 3.9 & 3.8 & 4.3 & 3.7 & 3.5 \\
  & &
  \hl\textbf{SparseDecoding}
  & \hl\textbf{5.31}
  & \hl{5.2} & \hl{5.3} & \hl{5.4} & \hl{4.7} & \hl{5.8} & \hl{5.7}
  & \hl{5.4} & \hl{5.6} & \hl{5.3} & \hl{6.0} & \hl{5.3} & \hl{5.5} \\
\hline
\end{tabular}
\end{table*}

\paragraph{Ablation over calibration datasets (Table~\ref{tab:ablation-calib-source}).}

We extend our comparison to two additional widely used calibration datasets, the Pile~\citep{gao2020pile} and RedPajama~\citep{weber2024redpajamaopendatasettraining}, evaluating Qwen3-14B and Qwen3-32B under 50\% unstructured and 2:4 semi-structured sparsity. SparseDecoding consistently achieves higher overall WritingBench scores than calibration with either corpus across both models and sparsity settings.

\section{Kernel Implementation Details}
\label{appendix:kernel}

\paragraph{Implementation.}
Our \(N{:}M\) SpMV kernel is implemented in Triton~\citep{tillet2019triton} and optimized for batch-one autoregressive decoding. We avoid explicitly staging the input vector in shared memory, where input-vector loads use the \texttt{.ca} cache policy\footnote{See the \href{https://docs.nvidia.com/cuda/parallel-thread-execution/index.html\#cache-operators}{NVIDIA PTX ISA documentation} for definitions of cache operators.}, so that activations reused across output rows can benefit from both L1 and L2 caching, while sparse weights and bitmask metadata use the \texttt{.cg} policy and are streamed primarily through L2 since they are primarily streamed and have limited temporal
reuse.
We compile a separate kernel for each \(N{:}M\) pattern, fixing \(N\) and \(M\) at compile time.
This fixes the number of nonzeros per group and the bitmask-decoding loop bounds at compile time, allowing the compiler to fully unroll the corresponding loops. 
Each nonzero position is recovered using a bit-scan operation that locates the least significant set bit. We also autotune the output tile size, number of warps, pipeline stages, and reduction strategy for each matrix shape and sparsity pattern. For configurations that benefit from additional parallelism, we split the reduction across multiple programs and combine the partial sums with \texttt{atomicAdd}. 

\end{document}